\documentclass{article} 
\usepackage{iclr2017_conference,times}
\usepackage{hyperref}
\usepackage{url}
\usepackage{graphicx}
\usepackage{caption}
\usepackage{subcaption}
\usepackage{amsmath}

\newtheorem{remark}{Remark}

\newcommand{\esm}[1]{\ensuremath{#1}}

\newcommand{\ms}[1]{\esm{\mathsf{#1}}}
\newcommand{\mi}[1]{\esm{\mathit{#1}}}

\newcommand\xbase{x'}

\newcommand\sparam{\alpha}

\newcommand\synteq{::=}

\newcommand\integratedgrads{\ms{IntegratedGrads}}

\newcommand\spearmannz{$\mi{spearman}_{\mi{nz}}$}

\title{A Note about: Local Explanation Methods for Deep Neural Networks Lack Sensitivity to Parameter Values}

\author{Mukund Sundararajan \& Ankur Taly \\
Google Inc.\\
Mountain View, CA 94043, USA
\\ \texttt{\{mukunds,ataly\}@google.com} \\ }

\begin{document}

\maketitle

\begin{abstract}

 Local explanation methods, also known as attribution methods, attribute a deep network's prediction to
  its input (cf.~\cite{BSHKHM10}). We respond to the claim from \cite{AGGK18} that local explanation
methods lack sensitivity, i.e., “DNNs with randomly-initialized
weights produce explanations that are both visually and quantitatively
similar to those produced by DNNs with learned weights.” 

Further investigation reveals that their findings are due to two choices in their analysis: (a) ignoring the signs of the attributions; and (b) for integrated gradients (IG), including pixels in their analysis that have zero attributions by choice of the baseline (an auxiliary input relative to which the attributions are computed). When both factors are accounted for, IG attributions for a random network and the actual network are uncorrelated. Our investigation also sheds light on how these issues affect visualizations, although we note that more work is needed to understand how viewers interpret the difference between the random and the actual attributions.
\end{abstract}

\section{Introduction}

Local explanation methods (also known as attribution methods) attribute a deep network's prediction
to its input (cf.~\cite{BSHKHM10,SVZ13,SGSK16,BMBMS16,SDBR14,LL17, STY17}). For instance, the network may perform an object recognition task.
It takes as input an image and predicts scores for classes that are synonymous with objects in the image. The attributions then assign importance scores
to the pixels.

We respond to the claim from \cite{AGGK18} that local explanation
methods lack sensitivity, i.e., “DNNs with randomly-initialized
weights produce explanations that are both visually and quantitatively
similar to those produced by DNNs with learned weights.”

In this response, we show that the two methods (Spearman correlation
and visual similarity) that~\cite{AGGK18} used to demonstrate this
insensitivity made certain choices due to which they saw this insensitivity.
To be clear, we do not show that local
explanation methods are trustworthy. Rather, we attempt to demonstrate that~\cite{AGGK18}
does not seem to show that they are not. Specifically, we only address
claims for one of the local explanation methods studied, i.e.,
Integrated Gradients, but the issues with the investigation possibly apply to other methods.
Likewise, we also only discuss results about MNIST, but the impact of these choices likely applies to other tasks too.

\section{Methodology of~\cite{AGGK18}}
\cite{AGGK18} studies sensitivity of an explanation method to network parameters
by randomizing the parameters starting from the topmost layer
and going all the way to the bottom.\footnote{They call this ``cascading randomization''
  of the network. Randomization involves reinitializing parameters with truncated normal distribution with mean $0$ and standard deviation $0.01$.} They then visually
and quantitatively assess the changes in the generated explanation
for a given set of inputs. The quantitative assessment involves
computing the spearman rank correlation between the explanation vector
from the original network and one with parameters randomized. If the correlation is high,
then the network is (supposedly) not sensitive to the parameters that were randomized, and therefore the importance
scores assigned by the attribution method are suspect. The explanation vectors are normalized by taking their absolute values.

\section{Sensitivity of Integrated Gradients}
We carefully examine the visualizations and the spearman rank
correlation for explanations produced using the Integrated Gradients
method~\cite{STY17}. Formally, the integrated gradient for the  $i^{th}$ base feature (e.g. a pixel) of an input $x$ and baseline
$\xbase$ is defined as:
\begin{equation}
\label{eq:intgrads}
\integratedgrads_i(x) \synteq (x_i-\xbase_i)\cdot\int_{\sparam=0}^{1} \tfrac{\partial F(\xbase + \sparam (x-\xbase))}{\partial x_i  }~d\sparam
\end{equation}
where
$\tfrac{\partial F(x)}{\partial x_i}$ is the gradient of
$F$ along the $i^{th}$ dimension at $x$.

It is claimed that these importance scores remain visually and
quantitatively unchanged as the the network is randomized. We argue that
these findings are due to certain artifacts of how the visualization
and spearman rank correlation is computed.

\subsection{Spearman rank correlation}
\begin{figure}
  \includegraphics[width=0.7\textwidth]{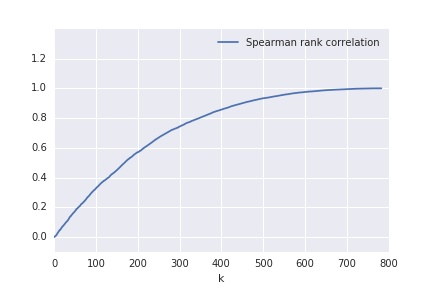}
  \caption{\textbf{Spearman rank correlation between two vectors as a function of the number of common indices where both vectors are zero.}
    The y-axis shows the spearman rank correlation value between two positive random vectors of size
    $784$ (which is the size of an MNIST input). The plot show how the correlation changes as the number of common
    indices where both vectors are zero changes increases from $0$ to $784$.}\label{fig:spearmanmask}  
\end{figure}

We begin by mentioning a property of spearman rank correlation.
Consider any two positive
vectors $a_1, a_2 \in R^n$ such that there exist $k$ common indices where $a_1$
and $a_2$ are both zero. The spearman rank correlation between $a_1$
and $a_2$ saturates to 1.0 as $k$ increases. Figure~\ref{fig:spearmanmask}
shows this empirically for two positive random vectors. Even when
$k$ is only 50\% of the indices, the spearman rank correlation is more
than $0.8$.

\cite{AGGK18} reports results on sensitivity of integrated gradients for a
3-hidden layer CNN trained over the MNIST dataset. Furthermore, the
integrated gradient vectors are normalized by taking absolute value
before computing spearman correlation.

Integrated gradients explain the network’s prediction for an input
relative to a certain baseline. In~\cite{AGGK18}, as~\cite{STY17} recommends, Integrated gradients
are computed using a black image as baseline. A mathematical property of the method is that
features (pixels in the case of image models) that have identical
values at the input and the baseline are guaranteed to receive zero
attribution.

Therefore, if a1 and a2 are integrated vectors from two different
networks for the same MNIST image and black baseline, then a1 and a2
would be identically zero for all pixels that are black in the
image. A typical MNIST image consists of a large number (nearly $80\%$)
of black pixels, and therefore by the property stated at the beginning of
this section, $\mi{abs}(a_1)$ and $\mi{abs}(a_2)$ would have a high spearman rank
correlation; here $\mi{abs}$ is the absolute value function. Note that this is
true regardless of the network architecture or parameters of the two
networks.

The above result explains why ~\cite{AGGK18} empirically find high spearman
correlation (nearly $1.0$) between (normalized) integrated gradients as the network
parameters are gradually randomized. We consider this high correlation
artifactual as it follows from the choice of baseline. A better
correlation metric would be to consider the spearman rank correlation
between the integrated gradient values of \emph{only} those pixels that change between the
baseline and input. When the baseline is black this amounts to
considering only those pixels that are nonzero in the input
image. We call this metric \spearmannz.

\begin{figure}
  \includegraphics[width=0.7\textwidth]{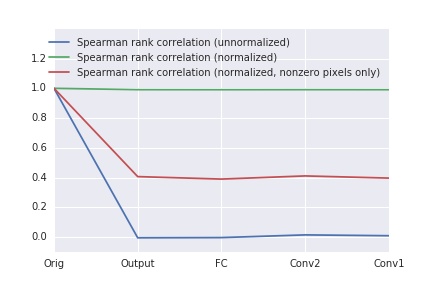}
  \caption{\textbf{Spearman rank correlation as the network is randomized.}
    The y-axis shows the correlation value. Each point on the x-axis
    corresponds to a network with paramaters upto that layer
    randomized. The blue plot shows the spearman rank correlation
    between the unnormalized integrated gradients from the original
    trained network and a randomized network. The green plot shows the
    correlation after normalizing the integrated gradient vectors, and
    the red plot shows the correlation after normalizing and only
    considering the pixels that are nonzero in the input, i.e., the
    \spearmannz\ correlation. All plots report the average correlation
    over $50$ randomly sampled MNIST images.}\label{fig:spearman}
\end{figure}

We now use the \spearmannz\ metric to compare (normalized) integrated
gradients computed for a trained network with that computed for a network
with randomized parameters. Figure~\ref{fig:spearman} (red plot)
shows the trend of the \spearmannz\ 
correlation as parameters of the network are randomized in a cascading manner
starting from the ``Output'' layer down to the first convolutional layer
(called ``Conv1'').  It is clear that the \spearmannz\ 
correlation drops sharply as soon as parameters in the ``Output'' layer are randomized.

\begin{remark}
  We notice in Figure~\ref{fig:spearman} that the \spearmannz\ correlation plateaus
  around $0.4$ after the ``Ouput'' layer is randomized. This can be explained as follows. Once the output layer
  is randomized, the gradient of the prediction with respect to the
  input pixels is effectively random. Now, why does the spearman correlation not go to zero?
  The mathematical form of integrated gradients includes scaling the integration of the gradients by
  the input, i.e., in Equation~\ref{eq:intgrads}, the term $(x_i-\xbase_i)$; recall that $\xbase_i$is zero because the baseline is black.
  That is, the mathematical form includes a non-random component that sets a bound on how low the correlations can go, i.e.,
  when the parameters are random, the integrated
  gradients important scores are effectively $\langle \mi{x}\rangle \times \langle \mi{random vector} \rangle$.
  We empirically find that the average \spearmannz\ correlation between two such
  integrated gradient vectors is $0.424$ ($99$\% CI [$0.413$, $0.437$]), which is what the correlation
  between the integrated gradients for the original network and those for the randomized network approximately
  converges to in Figure~\ref{fig:spearman}.
\end{remark}

\begin{remark}
We find that the normalization of the integrated gradients by taking
the absolute value is crucial for the findings in ~\cite{AGGK18}. In
particular we find that without the normalization, correlation under
the original spearman metric drops all the way to $0.0$ as parameters
of the network are randomized; see the blue plot in Figure~\ref{fig:spearman}.
\end{remark}


\subsection{Visualization}
\begin{figure}[!ht]
  \centering
  \includegraphics[width=0.5\textwidth]{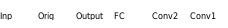}
  \includegraphics[width=0.5\textwidth]{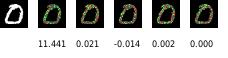}
  \includegraphics[width=0.5\textwidth]{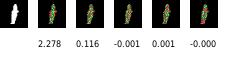}
  \includegraphics[width=0.5\textwidth]{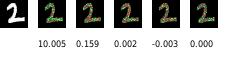}
  \includegraphics[width=0.5\textwidth]{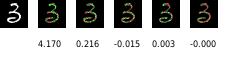}
  \includegraphics[width=0.5\textwidth]{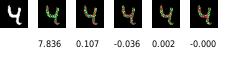}
  \includegraphics[width=0.5\textwidth]{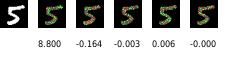}
  \includegraphics[width=0.5\textwidth]{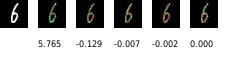}
  \includegraphics[width=0.5\textwidth]{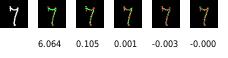}
  \includegraphics[width=0.5\textwidth]{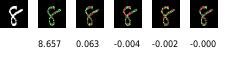}
  \includegraphics[width=0.5\textwidth]{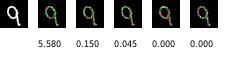}
  \caption{\textbf{Visualizations of integrated gradients as the network is randomized.}
    Each row shows a different MNIST image. The first and second columns show the
    image and the visualization of integrated gradients from the original trained
    network. The following columns show visualization of integrated gradients (for the same label as the original image) as
    parameters in each layer of the network are successively randomized in a cascading
    manner. The number below each visualization is the sum of the raw integrated gradient
    scores of all pixels. It approximates the difference between the logit score for the input and the baseline (i.e., black image)}\label{fig:vis}
\end{figure}

\cite{AGGK18} show that visualizations of normalized integrated gradients
are largely unchanged as parameters of the network are randomized.
We notice that the blame for this falls on the
normalization, i.e., taking absolute values. In particular, doing this conflates pixels
that receive positive attribution with those that receive negative
attribution.  A more faithful visualization is one that preserves the
sign, and shows pixels with positive and negative attribution on
separate channels.

Figure~\ref{fig:vis} shows such visualizations of integrated gradients for a few
MNIST images. It also shows how the visualizations change as
parameters of the network are gradually randomized. (All visualizations are
for the same label as that of the original image.)
It is clear that the
sign of the integrated gradient value of a pixel is highly sensitive to the
parameters of the network. For instance, notice how different parts of
the number ``8'' flip from positive to negative attribution as the network is
randomized. Notice also that the attributions for the original unrandomized network appear to highlight
entire strokes whereas the attributions for the randomized networks appear disjointed.

Furthermore, while the visualizations in Figure~\ref{fig:vis} preserve the sign of the
integrated gradients vectors, they do not reflect its magnitude. Each
visualization is obtained by scaling the min and max values to $0$ and
$255$ respectively, and therefore destroys the magnitude information.
A mathematical property of integrated gradients is that the sum of the
integrated gradient scores of all features adds up to the prediction score
at the input minus the prediction
score at the baseline. As parameters of the network are randomized,
the final prediction score for the original label drops significantly,
and so does the sum of the integrated gradients across all pixels.
We verify this empirically in Figure~\ref{fig:vis}. Notice that
the sum of the integrated gradient values across all pixels approaches
$0$ as parameters of the network are randomized.

That said, we do agree that further work needs to be done on careful visualizations for attribution. For image-based use-cases, humans consume the visualizations and not the numbers and if the visualizations are not carefully done, we risk losing relevant information contained in attribution scores.

\section{Acknowledgements} We thank Julius Adebayo, Been Kim, and Raz Mathias for helpful discussions.

\bibliography{rebuttal}
\bibliographystyle{iclr2017_conference}
\end{document}